\documentclass[letterpaper, 10 pt, conference]{ieeeconf}

\IEEEoverridecommandlockouts
\usepackage{times}
\usepackage[bookmarks=true,hidelinks]{hyperref}
\usepackage{tabularx}
\usepackage{graphicx}
\usepackage{amsmath}
\usepackage{epsfig}
\usepackage{subfig}
\usepackage{cite}
\usepackage{float}
\usepackage{ifthen}
\usepackage{xcolor}
\usepackage{ntheorem}
\theoremseparator{:}
\usepackage{soul}
\usepackage{colortbl}
\usepackage{booktabs}
\let\labelindent\relax 
\usepackage{enumitem}
\usepackage{tikz}
\usetikzlibrary{arrows.meta,positioning,calc}

\usepackage[font=footnotesize]{caption}
\setlist[itemize]{leftmargin=*}

\newcommand{\fig}[1]{Fig.~\ref{#1}}

\title{\LARGE \bf Correcting Persistent Failures in Point-Cloud Grasping Policies using Human Demonstrations in Virtual Reality}

\author{Achyuthan Unni Krishnan$^{*}$, Karthik Rengarajan$^{*}$, Bharath Narasimhan, and Chris von Csefalvay%
\thanks{$^{*}$Achyuthan Unni Krishnan and Karthik Rengarajan contributed equally to this work.}%
}

\begin{document}

\maketitle
\thispagestyle{empty}
\pagestyle{empty}

\begin{abstract}
Robots in homes, warehouses and farms must pick up objects they have never seen, usually without a 3D model or a category label. A grasping policy that works directly on the point cloud from a depth sensor needs only the geometry the robot observes, so it applies to any new object. Offline grasp generators such as Contact-GraspNet work this way, but they rank grasps by learned plausibility rather than by whether the grasp actually lifts the object. Reinforcement learning (RL) can optimize for that directly and make a point-cloud grasping policy much better on average, yet it leaves specific grasps persistently unsolved. We present a human-in-the-loop failure loop that targets those grasps: situations an RL policy keeps failing are mined automatically, sent to a virtual-reality application where a person picks the object up with their own hands, and returned to training as two grasp points. The person contributes only where to grasp; the robot chooses its own approach. In simulation, RL raises success on unseen objects from 55\% with Contact-GraspNet to 83\%. Adding 58 human demonstrations nearly doubles success on the hard cases RL kept failing, from 24\% to 45\%, and recovers performance that RL alone loses on similar cases that were never demonstrated, without reducing success on unseen objects.
\end{abstract}

\section{Introduction}\label{sec:intro}

Grasping policies operating in point-cloud space work without a category label or classifier; instead, they operate on learned geometric features that tend to lead to success. Learning to grasp from point clouds has two complementary sources of skill. Grasp generators trained on large offline datasets propose plausible grasps for new objects~\cite{sundermeyer2021contact}, but high scores do not guarantee that a grasp survives physical execution. Reinforcement learning (RL) in simulation optimizes physical success directly~\cite{schulman2017ppo}, but it depends on exploration and averages over everything it sees, so a policy can improve overall while particular object-pose situations remain unsolved. For example, in our experiments, RL raises success on unseen objects by 18--25 percentage points over the policy it starts from (24.5 flat, 17.5 tilted), but ends 10--13 points below it on hard cases (\fig{fig:tradeoff}).

Demonstrations are a natural remedy, and combining them with RL is well established~\cite{vecerik2017ddpgfd,nair2018overcoming,rajeswaran2018learning}. Two practical questions remain for grasping. First, \emph{which} situations deserve a person's limited time? Collecting demonstrations uniformly spends effort on grasps the policy already performs. Second, \emph{what} should a person demonstrate? A human hand and a parallel-jaw gripper have different kinematics, so copying the whole hand motion imposes constraints the robot does not share. Robot motion to the grasp pose is outside the scope of this work and is assumed to be handled by a standard motion planner.

We address both questions with a \emph{failure loop} (\fig{fig:overview}). Training records every failed grasp and replays it later; situations that keep failing are exported to a VR application where a person grasps and lifts the object. The two contact points return to training and answer \emph{where} to grasp, and the robot remains free to choose its approach. An automatic check executes each demonstration with the robot's own gripper to confirm the validity of the demonstration. Our contributions are:

\begin{figure}[t]
    \centering
    \includegraphics[width=\linewidth]{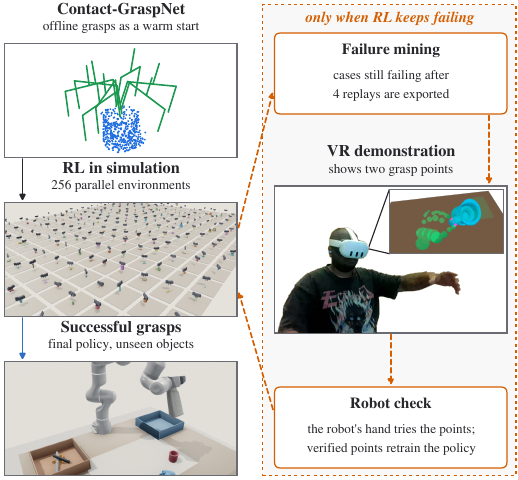}
    \caption{The failure loop. RL trains a policy warm-started from Contact-GraspNet in 256 parallel environments. Grasps it keeps failing (orange) go to VR, where a person shows two grasp points; points verified with the robot's hand retrain the policy from the same warm start. Bottom: the final policy on unseen objects.}
    \label{fig:overview}
    \vspace{-2ex}
\end{figure}

\begin{itemize}
    \item A failure loop that sends persistently failing grasps to VR and returns contact points only, verified by execution with the robot's hand.
    \item A training rule in which verified demonstrations are executed and imitated only after they lift the object, with imitation fading out so that RL keeps the final policy.
    \item Evaluation on real-scale objects showing that RL improves on Contact-GraspNet for unseen objects by 28--30 percentage points but loses hard cases that the warm start could solve, and that demonstrations recover them, including undemonstrated poses of demonstrated objects, with no measurable loss on test objects. A warm-start control and a split between demonstrated and hidden (never demonstrated) hard cases show the impact of demonstrations.
\end{itemize}
Code for the training pipeline and the VR demonstration application is available at \url{https://github.com/ukachyuthan/unitree-grasping} and \url{https://github.com/ukachyuthan/vr-grasping-project}.

\section{Related Work}\label{sec:related}

\emph{Grasping objects unlike the training data.} Analytic grasp planning needs an object model and accurate pose, so learned methods that map sensor data directly to grasps have become standard~\cite{bohg2014survey}. From RGB and RGB-D images, convolutional networks detect grasp rectangles~\cite{lenz2015deep,redmon2015realtime}, and large self-supervised robot datasets learn grasp success directly~\cite{pinto2016supersizing,levine2018handeye}. From depth and point clouds, grasp-quality networks are trained on synthetic data labelled by analytic metrics~\cite{mahler2017dexnet2,mahler2019ambidextrous,morrison2018closing}, and 6-DoF generators propose full gripper poses for unseen objects~\cite{tenpas2017gpd,mousavian2019graspnet,sundermeyer2021contact}, supported by large grasp datasets~\cite{fang2020graspnet1b,eppner2021acronym}. In both modalities, success drops when objects, viewpoints, or scenes differ from the training distribution, and a grasp that scores highly need not survive physical execution. Bridging the gap from simulation to the real world requires deliberate randomization of the training data~\cite{tobin2017domain}. We use Contact-GraspNet as both a baseline and a warm start; on our held-out objects its top grasp lifts the object in about half of the trials.

\emph{Reinforcement learning for grasping.} RL optimizes physical grasp success rather than a proxy score. Vision-based grasping with off-policy RL has been compared in simulation~\cite{quillen2018deep} and scaled to real robots with hundreds of thousands of attempts~\cite{kalashnikov2018qtopt}. Such policies depend on exploration and optimize average success, so specific object--pose situations can remain unsolved while overall success rises. We start RL from an offline generator rather than from scratch, train on point clouds of real and procedurally morphed objects in simulation, and show that RL improves on the generator for objects never seen in training. TI-0.1~\cite{tutor2026ti01} takes a complementary, hardware-driven route: agents rebuild a task in simulation from a few teleoperated demonstrations and telemetry, RL trains there, and hardware rollouts refine the reconstruction. There, failures correct the simulator; in our loop, they go to a person, who corrects the policy.

\emph{Demonstrations and human input.} Demonstrations are a common remedy for hard exploration, through replay buffers~\cite{hester2018dqfd,vecerik2017ddpgfd}, auxiliary imitation losses~\cite{nair2018overcoming}, or pretraining with augmented policy gradients~\cite{rajeswaran2018learning}. Interactive imitation asks a person for labels on states the learner actually visits~\cite{ross2011dagger,kelly2019hgdagger}, and VR and crowdsourced teleoperation make demonstration collection practical~\cite{zhang2018vr,mandlekar2018roboturk}. Our work combines these ideas for grasping in three ways. First, demonstrations are requested only for situations the RL policy keeps failing, much as hard-example mining focuses training on the examples a detector gets wrong~\cite{shrivastava2016ohem}. Second, a person shows only where to grasp, as two contact points, which relates to work that learns grasp contacts from humans~\cite{brahmbhatt2019contactdb}; the robot keeps its own choice of approach, so human and gripper kinematics need not match. Third, each demonstration is verified with the robot's hand before training uses it. Human input is also used during execution in shared-autonomy teleoperation, where an operator guides a mobile humanoid through VR or gaming interfaces while the robot handles collision avoidance and base--arm coordination~\cite{boguslavskii2023shared}, or shares control of torso--arm coordination with the operator~\cite{boguslavskii2025human}. These systems keep a person in the control loop throughout the task, whereas here the person contributes grasp points for selected failures and the trained policy then acts autonomously.

Offline grasp generators propose plausible grasps for new objects but score them by a proxy, so a highly ranked grasp may still fail when executed. Their scores are learned from proxy labels on static geometry~\cite{mahler2017dexnet2,eppner2021acronym}, so they do not account for contact dynamics during closing, mass and friction during the lift, partial observations, or the kinematics of a different gripper. Analytic grasp-quality metrics of this kind are known to predict physical grasp success poorly~\cite{rubert2017relevance}. We use such a generator only as a warm start and let RL optimize physical success. RL in turn raises average success but leaves specific object--pose situations unsolved. We mine those failures automatically and request demonstrations only for them to improve upon the RL policy's blind spots.

\section{Proposed Method}\label{sec:proposed}

\subsection{Grasp Policy}
The policy maps a point cloud of 128 object points to one grasp in a single decision. The points are centered on their noisy centroid with world-aligned axes, and the observation contains no object pose. During training the points are sampled from the whole object surface, so the policy assumes a view that covers most of the object. Sensor noise includes jitter (up to 5\,mm), dropout (up to 10\% of points), and outliers. The action has seven dimensions: two contact points $c_1, c_2 \in [-12,12]^3$\,cm and an approach angle $\psi \in [-90^\circ, 90^\circ]$ about the jaw axis. Swapping the fingers ($c_1 \leftrightarrow c_2$, $\psi \to -\psi$) gives the same grasp. Actor and critic use separate PointNet encoders~\cite{qi2017pointnet} with a 128-dimensional embedding followed by a $128$--$64$ MLP.

We focus on grasp selection rather than arm motion, so a floating Franka hand executes each grasp through a scripted sequence: settle, insert, close, lift, and hold. A grasp succeeds only if the object rises by at least 5\,cm with the fingers in contact. Shaping rewards never count as success.

\begin{figure*}[t]
    \centering
    \includegraphics[width=\linewidth]{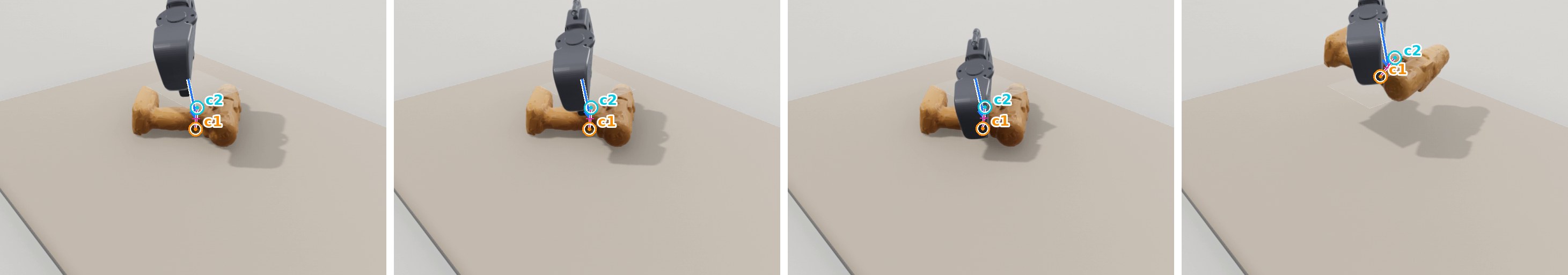}
    \vspace{-3ex}
    \caption{A simulated grasp by the floating Franka hand (settle, insert, close, lift). RL + demos acts on a held-out power drill; $c_1$ and $c_2$ are its predicted contacts.}
    \label{fig:sim}
    \vspace{-2ex}
\end{figure*}

\subsection{Warm Start and RL}
The actor is first trained by behavior cloning on grasps generated offline by Contact-GraspNet~\cite{sundermeyer2021contact}, using the distance to the nearest labelled grasp as the loss. RL then runs one-step PPO~\cite{schulman2017ppo} in Isaac Lab~\cite{mittal2023orbit}. The reward is
\begin{equation}
    r = r_{\text{lift}} + 0.3\,r_{\text{contact}} + 0.4\,r_{\text{surface}} + r_{\text{graspnet}},
\end{equation}
where $r_{\text{lift}}$ is the lift height as a fraction of the 5\,cm target, zero without finger contact, and the remaining terms reward finger coverage of the object, contacts close to its surface, and proximity to a high-quality Contact-GraspNet grasp.

\subsection{Failure Mining}
During training, each failed grasp enters a replay queue with its exact object and pose, and 10\% of episodes re-run queued failures. A replay that succeeds removes the entry. Four failed replays export it as a \emph{case} containing the mesh, pose, point cloud, and failed attempts. Because replay mining can export lucky failures, we retest every exported case on the final policies of all RL seeds before showing it to a person, and keep only those that rarely lift.

\subsection{VR Demonstration}
Cases are presented on a Meta Quest 3 through a WebXR application built on tracked hands. The object appears at its recorded pose on a virtual table, and the person grasps and lifts it, holding it for 0.5\,s. The application stores the two opposing fingertip contacts in the object frame. It also stores the hand trajectory, which we do not use. If a person judges a case physically impossible, they can mark it as such and move on.


Demonstrations contribute \emph{points only}. Because the hand and gripper differ, we neither judge nor imitate the person's hand orientation.

\subsection{Robot Check}
A live check server executes each recorded pair of points with the robot's hand while the person is still in the headset. It searches approach angles from $-80^\circ$ to $80^\circ$ in $20^\circ$ steps. If some angle lifts the object, it repeats that grasp four times under execution noise of 1\,cm and $5^\circ$. A demonstration is \emph{robust} when at least half of the repeats lift, and only robust demonstrations enter training. The person sees the outcome immediately and can try again.

\subsection{Training with Demonstrations}
\label{sec:training}
RL + demos starts from the same warm start as RL alone and uses the same recipe. On 10\% of episodes, a demonstrated case is replayed at its exact pose. Half of those episodes execute the demonstrated grasp at the approach angle found by the check, and the other half execute the policy's own grasp as usual. Executed demonstrations train the critic but are excluded from the policy-gradient term. The actor imitates a demonstration only on an executed episode where it lifted:
\begin{equation}
    \mathcal{L} = \mathcal{L}_{\text{PPO}} + \lambda(t)\,\frac{1}{|\mathcal{D}_t|}\sum_{i \in \mathcal{D}_t} \min\!\big(\lVert a_i - a_i^{*} \rVert,\ \lVert a_i - \tilde a_i^{*} \rVert\big),
\end{equation}
where $\mathcal{D}_t$ is the set of those episodes, $a_i$ and $a_i^{*}$ are the policy mean and the demonstration restricted to the two contact points, $\tilde a_i^{*}$ swaps the fingers, and $\lambda(t)=\max(0, 1-(t-1)/40)$ fades imitation to zero by iteration 41. If $\mathcal{D}_t$ is empty, the imitation term is zero and the update is plain PPO. This did not occur in practice: at every logged iteration (every tenth) of the imitation phase, at least 105 executed demonstrations lifted across all three seeds, about 26 per minibatch. The approach angle is never imitated. For the remaining 110 iterations the demonstrated cases keep appearing, but RL alone decides what to do with them.

This design follows from an earlier experiment on a different object set, in which imitating demonstrations on every demonstration episode pulled the policy away from grasps that already worked and lowered hard-case performance.

\begin{figure*}[t]
    \centering
    \includegraphics[width=\linewidth]{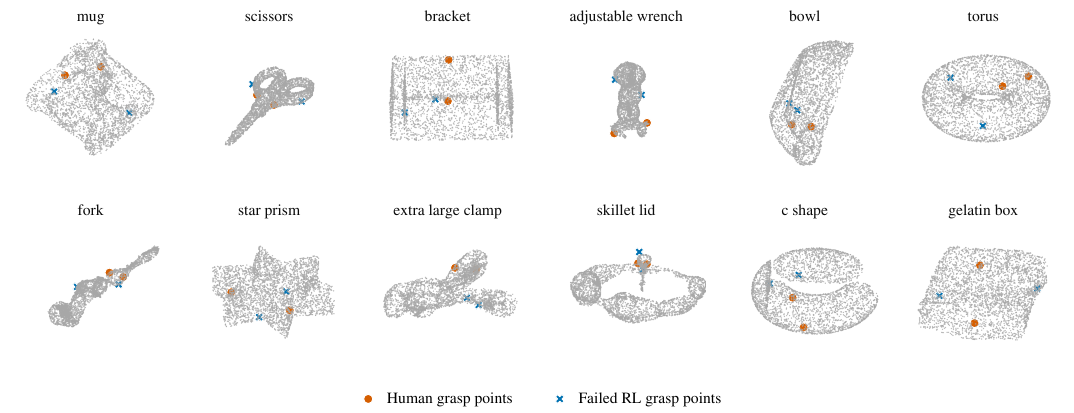}
    \vspace{-3ex}
    \caption{Human grasp points (orange) and a failed RL grasp (blue) on one demonstrated case per family, chosen where RL + demos improved most, so not a representative sample. Gray: object surface points.}
    \label{fig:demos}
    \vspace{-2ex}
\end{figure*}

\section{Experimental Setup}\label{sec:exp}

\subsection{Objects and Simulation}
Objects are YCB models~\cite{calli2015ycb} and procedural shapes such as brackets, tori, and star prisms. We use real-scale meshes up to 30\,cm whose concavities are preserved, so bowls and mugs stay hollow. Training uses 451 instances from 57 families, and testing uses 24 separate YCB families. Fingertips may touch and slide along the table. Simulation runs at 60\,Hz in Isaac Lab~2.3.2 with Isaac Sim~5.1.

\subsection{Training Recipe}
Every condition uses the settings in Table~\ref{tab:params}, for about 614,000 grasps per seed. The demonstration settings were fixed before any RL + demos result on this object set was seen.

\begin{table}[t]
    \centering
    \caption{Training settings. The last group applies to RL + demos only.}
    \label{tab:params}
    \begin{tabularx}{\linewidth}{X l}
    \toprule
    \textbf{Setting} & \textbf{Value} \\
    \midrule
    Environments $\times$ decisions per iteration & $256 \times 16$ = 4,096 \\
    Iterations; checkpoint interval & 150; 25 \\
    Learning rate; clip; entropy coefficient & $3\times10^{-4}$; 0.2; 0.01 \\
    Epochs; minibatches & 5; 4 \\
    Action noise at start; floor & 0.3; 0.05 \\
    Training execution noise & 1\,cm, $5^\circ$ \\
    Objects spawned tilted during training & 10\% \\
    Failure replay share of episodes & 10\% \\
    \midrule
    Demonstration episodes & 10\% \\
    Demonstration grasps executed & 50\% of those \\
    Imitation weight & 1 $\to$ 0 over 40 iterations \\
    \bottomrule
    \end{tabularx}
    \vspace{-2ex}
\end{table}

\subsection{Hard Cases and Demonstrations}
\begin{figure}[t]
    \centering
    \includegraphics[width=\linewidth]{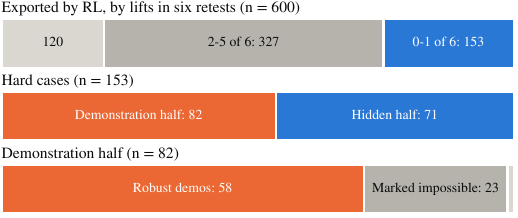}
    \vspace{-3ex}
    \caption{From exported failures to training demonstrations. One demonstration-half case was skipped.}
    \label{fig:funnel}
    \vspace{-2ex}
\end{figure}

Three RL-alone seeds exported 600 cases. Each policy retried each case twice without execution noise. Of the resulting six tries, 120 cases lifted every time, 327 lifted two to five times, and 153 lifted at most once. Those 153 are our hard cases (\fig{fig:funnel}). We split them by a stratified random draw within object family into a demonstration half (82) and a hidden half (71) that the VR application never shows.

One demonstrator, an author, worked through the demonstration half in a single 32-minute session with 113 attempts. This produced 58 robust demonstrations, 23 cases marked impossible, and one skipped case. The cases marked impossible are excluded from the demonstrated-case score, leaving 59 cases. All 71 hidden cases are scored, including any that may be impossible.

\subsection{Conditions and Metrics}
We compare \emph{RL alone} with \emph{RL + demos} over three seeds each, tested every 25 iterations. As references we test Contact-GraspNet~\cite{sundermeyer2021contact} (the top-ranked grasp, and the best of the top ten after filtering) and the \emph{warm start} before any RL. All tests run without execution noise:
\begin{itemize}
    \item \emph{Test objects}: 200 episodes on the test families, flat (seed 31) and tilted (seed 32). A few episodes per run produce no result and are not counted.
    \item \emph{Hard cases}: two attempts on each case at its exact pose, scored on the demonstrated and hidden halves separately.
\end{itemize}
The primary comparison is at iteration 150. We pool each case over seeds and repeats, then apply a two-sided sign-flip permutation test on the per-case differences, with a bootstrap 95\% interval over cases. Test objects are compared in the same way, with the object as the unit.

\section{Results And Discussion}\label{sec:res}

\begin{figure}[t]
    \centering
    \includegraphics[width=\linewidth]{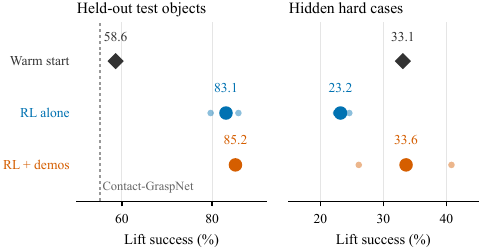}
    \caption{RL + demos is the only policy strong on both test objects and hidden (never demonstrated) hard cases. Iteration 150; large markers are three-seed means, small markers individual seeds; dashed line, Contact-GraspNet. Hard cases were selected as RL-alone failures (Sec.~\ref{sec:limits}).}
    \label{fig:tradeoff}
    \vspace{-2ex}
\end{figure}

\begin{table}[t]
    \centering
    \caption{Lift success (\%) without execution noise; RL rows are three-seed means at iteration 150. Hard cases were selected as RL-alone failures, which favors the other rows (Sec.~\ref{sec:limits}).}
    \label{tab:main}
    \begin{tabularx}{\linewidth}{X c c c c}
    \toprule
    & \multicolumn{2}{c}{\textbf{Test objects}} & \multicolumn{2}{c}{\textbf{Hard cases}} \\
    \cmidrule(lr){2-3}\cmidrule(lr){4-5}
    \textbf{Method} & \textbf{Flat} & \textbf{Tilted} & \textbf{Demo.} & \textbf{Hidden} \\
    \midrule
    Contact-GraspNet, top-1 & 55.1 & 50.3 & -- & -- \\
    Contact-GraspNet, filter & 64.3 & 49.2 & -- & -- \\
    Warm start, before RL & 58.6 & 62.7 & 37.3 & 33.1 \\
    RL alone & 83.1 & 80.2 & 24.0 & 23.2 \\
    \textbf{RL + demos} & \textbf{85.2} & \textbf{80.9} & \textbf{45.2} & \textbf{33.6} \\
    \bottomrule
    \end{tabularx}
    \vspace{-2ex}
\end{table}

\begin{figure*}[t]
    \centering
    \includegraphics[width=\linewidth]{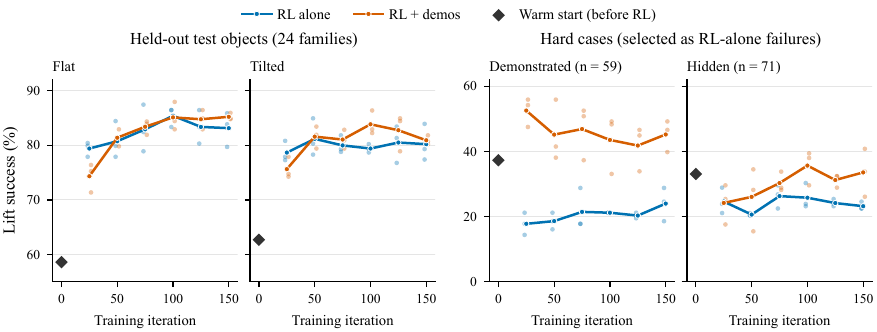}
    \vspace{-3ex}
    \caption{Success during training: three-seed means (lines) and individual seeds (dots).}
    \label{fig:curves}
    \vspace{-2ex}
\end{figure*}

\subsection{RL Improves on Contact-GraspNet}
Table~\ref{tab:main}, \fig{fig:tradeoff}, and \fig{fig:curves} summarize the results. On the 24 held-out test families, Contact-GraspNet's top-ranked grasp lifts the object in 55.1\% of flat and 50.3\% of tilted episodes; filtering its top ten grasps for the hand's reach raises the flat score to 64.3\%. The warm start, which imitates these grasps, scores 58.6\% and 62.7\%. RL from this warm start reaches 83.1\% flat and 80.2\% tilted, about 20 points above the warm start and 19--31 points above Contact-GraspNet. Optimizing physical success therefore generalizes to objects never seen in training. On the hard cases, however, RL alone ends below the warm start (24.0\% vs. 37.3\% demonstrated, 23.2\% vs. 33.1\% hidden), partly because these cases were selected as its failures.

\subsection{Demonstrations Improve Where RL Fails}
\begin{figure}[t]
    \centering
    \includegraphics[width=\linewidth]{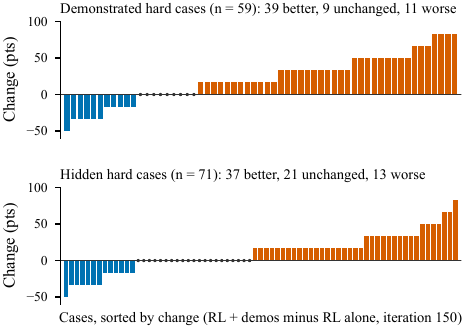}
    \vspace{-3ex}
    \caption{Per-case change from RL alone to RL + demos at iteration 150. Each bar is one hard case (three seeds $\times$ two attempts).}
    \label{fig:percase}
    \vspace{-2ex}
\end{figure}

On demonstrated hard cases, RL + demos reaches 45.2\% against 24.0\% for RL alone, a difference of $+21.2$ points (Table~\ref{tab:stats}). It stays above RL alone at every checkpoint, and 39 of 59 cases improve (\fig{fig:percase}). Because these are the demonstrated cases, part of the gain may reflect memorization of exact situations.

The hidden cases were never demonstrated. RL + demos reaches 33.6\% against 23.2\%, a difference of $+10.3$ points, with 37 cases improving and 13 worsening. The two conditions are level at iteration 25, and RL + demos leads at each later checkpoint. Pooling iterations 100--150 gives a similar $+9.1$ points.

Most hidden cases are demonstrated objects in conditions that were not demonstrated. On the 44 hidden cases whose exact object instance also appears among the demonstrations, but at another pose, RL + demos scores 32.6\% against 21.6\% for RL alone. On the 25 that use another instance of a demonstrated family, it scores 34.0\% against 24.7\%. Demonstrations on an object carry over to new poses of that object and to its variants. The two hidden cases from families with no demonstrations are too few to evaluate, and on the separate held-out test families, which had no demonstrations, RL + demos and RL alone perform the same (Sec.~\ref{sec:noloss}).

\begin{table}[t]
    \centering
    \caption{Mean differences at iteration 150 in percentage points, with 95\% bootstrap intervals: per case on hard cases, and per object on test families.}
    \label{tab:stats}
    \begin{tabularx}{\linewidth}{X c c}
    \toprule
    \textbf{Comparison} & \textbf{$\Delta$ (pts)} & \textbf{95\% CI} \\
    \midrule
    \multicolumn{3}{l}{\emph{Demonstrated cases ($n = 59$)}} \\
    RL + demos vs. RL & $+$21.2 & [$+$13.0, $+$29.4] \\
    RL + demos vs. warm & $+$7.9 & [$-$4.2, $+$19.8] \\
    RL vs. warm & $-$13.3 & [$-$24.6, $-$2.5] \\
    \midrule
    \multicolumn{3}{l}{\emph{Hidden cases ($n = 71$)}} \\
    RL + demos vs. RL & $+$10.3 & [$+$4.5, $+$16.2] \\
    RL + demos vs. warm & $+$0.5 & [$-$8.9, $+$9.9] \\
    RL vs. warm & $-$9.9 & [$-$18.8, $-$0.9] \\
    \midrule
    \multicolumn{3}{l}{\emph{Test objects, per object ($n = 24$)}} \\
    RL + demos vs. RL, flat & $+$1.2 & [$-$2.2, $+$5.5] \\
    RL + demos vs. RL, tilted & $+$0.2 & [$-$3.3, $+$3.8] \\
    \bottomrule
    \end{tabularx}
    \vspace{-2ex}
\end{table}

\subsection{Comparison With the Warm Start}
Before RL, the warm start already scores 33.1\% on hidden cases, equal to RL + demos. Its test-object scores, however, are far lower (58.6\% and 62.7\%). RL alone raises test-object success by about 20 points but ends about 10 points lower on hidden cases, and this difference is partly inflated by case selection. RL + demos makes the same test-object gain without that loss. On hidden cases, RL + demos therefore \emph{retains} competence that RL alone gives up; it does not exceed the warm start. On demonstrated cases, RL + demos is nominally 7.9 points above the warm start, but this difference is not significant.

Taken together, the warm start keeps hard-case performance but not test-object performance, RL alone does the reverse, and RL + demos is the only policy that holds both.

\begin{figure}[t]
    \centering
    \includegraphics[width=\linewidth]{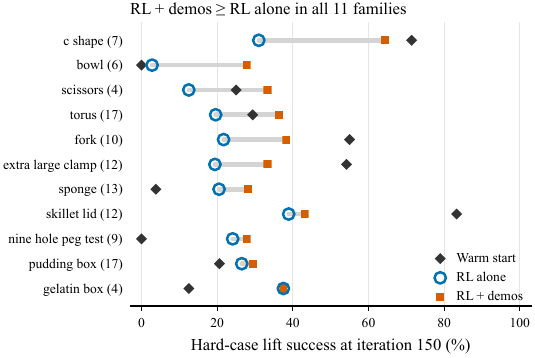}
    \vspace{-3ex}
    \caption{Hard-case success by family at iteration 150 (families with at least four cases; count in parentheses), demonstrated and hidden cases pooled.}
    \label{fig:families}
    \vspace{-2ex}
\end{figure}

\subsection{Differences Between Object Families}
\fig{fig:families} shows that the pattern is not uniform. RL + demos scores at least as high as RL alone in all eleven families with four or more scored cases, but the warm start splits them. On the skillet lid, clamp, fork, and C-shape families the warm start exceeds RL + demos, so both RL conditions lose part of what it could do there. On the bowl, nine-hole peg test, and sponge families the warm start is at or near zero, and RL + demos reaches 28\% in each against 3\%, 24\%, and 21\% for RL alone. These families have few cases each and we did not test them separately, so we leave as a hypothesis that demonstrations matter most where the offline grasp generator offers little.

\begin{figure*}[t]
    \centering
    \includegraphics[width=\linewidth]{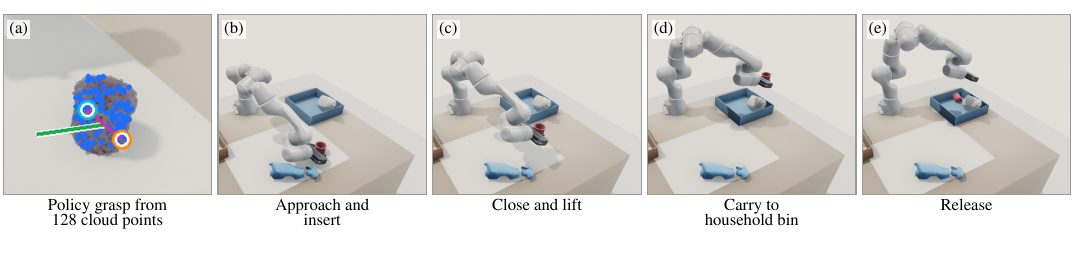}
    \caption{One sorting step by the Franka arm with the RL + demos policy. (a)~The policy receives 128 points of the cup's segment (blue; gray: the rest of the observed segment) and outputs two contacts (orange, cyan) and an approach (green). (b)--(e)~The arm inserts along the approach, closes, lifts, and places the cup in the household bin, on the first try. The cup is a test object not seen in training.}
    \label{fig:arm_sequence}
\end{figure*}

\subsection{No Loss on Test Objects}\label{sec:noloss}
On held-out families, RL + demos scores 85.2\% flat and 80.9\% tilted against 83.1\% and 80.2\% for RL alone, and the per-object differences are not significant (Table~\ref{tab:stats}). Both RL conditions score above the Contact-GraspNet references and the warm start.

\subsection{Sorting with a Robot Arm}\label{sec:sorting}
To test the policies outside the floating-hand setting, a Franka arm sorts six objects on a table into two bins by category. Six simulated depth cameras are fused and split into objects by the simulator's labels, standing in for an instance segmenter; each policy receives 128 points of an object's segment, as in training. The arm executes the grasp with multi-seed inverse kinematics, taking the mirrored approach ($-\psi$, same contacts) when the policy's approach is out of reach, and objects are free to move during the approach. Each object gets up to three tries: the policy's mean grasp, then samples. Contact-GraspNet uses its labelled grasps at the object's true pose, which the policies do not receive, trying the next reachable grasp of its filtered top ten each time. The scenes are \emph{fruit} (two of six are test objects), \emph{workbench} (six test objects: knife, clamp, marker, spray and bleach bottles, cup), and \emph{hard} (six hidden-case families at the yaw where RL alone failed). All methods see the same ten layouts per scene; each RL condition uses one seed.

\begin{table}[t]
    \centering
    \caption{Franka-arm sorting: objects in the correct bin within three tries, and lifted on the first try, out of 60 (ten layouts of six objects). One seed per RL condition; best per column in bold.}
    \label{tab:sorting}
    \begin{tabularx}{\linewidth}{X c c c c c c}
    \toprule
    & \multicolumn{2}{c}{\textbf{Fruit}} & \multicolumn{2}{c}{\textbf{Workbench}} & \multicolumn{2}{c}{\textbf{Hard}} \\
    \cmidrule(lr){2-3}\cmidrule(lr){4-5}\cmidrule(lr){6-7}
    \textbf{Method} & \textbf{Sorted} & \textbf{1st} & \textbf{Sorted} & \textbf{1st} & \textbf{Sorted} & \textbf{1st} \\
    \midrule
    Contact-GraspNet & 56 & 51 & 26 & 26 & 20 & 20 \\
    RL alone & 59 & 54 & 47 & 38 & \textbf{32} & \textbf{22} \\
    RL + demos & \textbf{60} & \textbf{57} & \textbf{50} & \textbf{47} & 28 & 20 \\
    \bottomrule
    \end{tabularx}
    \vspace{-2ex}
\end{table}

Table~\ref{tab:sorting} shows the results, and \fig{fig:arm_sequence} shows one sorting step on an unseen object. On the workbench objects, which no policy saw in training, both RL policies sort nearly twice as many objects as Contact-GraspNet (50 and 47 vs.\ 26; paired exact tests p $<$ 0.001 for each). On the hard scene, both RL policies also sort more (28 and 32 vs.\ 20; p = 0.02 and 0.004). Fruit is near the ceiling for all methods. RL + demos and RL alone do not differ significantly in any scene (workbench first try 47 vs.\ 38, p = 0.14; hard 28 vs.\ 32, p = 0.42). The hard-case gain of Sec.~\ref{sec:res} does not appear in this setting, where meshes, observations, and execution all differ from training and only one seed per condition is tested. Success also depended on seeing most of each object: in a preliminary run with three cameras on one side, three of six workbench objects were sorted, against five with six cameras.

\subsection{No Evidence of Faster Learning}
We expected demonstrations to reduce search and speed up learning. The data do not show this. At iteration 25 RL + demos is behind RL alone on test objects (74.3\% vs. 79.4\% flat, 75.6\% vs. 78.7\% tilted), and it is level by iteration 50. RL alone already approaches its final score by the first checkpoint, so our 25-iteration resolution cannot detect finer early differences, but it gives no support for a speed-up.

\subsection{What Did the Human Add?}
Table~\ref{tab:sources} compares sources of grasp points on the 82 demonstrated-half cases. In a single 32-minute session with 113 attempts, about 1.4 per case, the human produced robust grasp points for 58 cases, each confirmed by the noisy check above. This matches what random search reaches only with 32 attempts per case (54 cases; discordant cases 16 vs.\ 12, p = 0.57) and is roughly twice what two random attempts achieve (26--28). The human also supplies grasps that an offline generator misses: 17 cases were solved by the human but by none of Contact-GraspNet's top 40 grasps, and six cases were solved by the human and by neither automated source. Human points therefore cover hard cases efficiently and extend beyond the generator's proposals, and Section~\ref{sec:res} shows that they translate into measurable gains when used as training data.

\begin{table}[t]
    \centering
    \caption{Demonstrated-half cases (n = 82) solved by each source of grasp points; 26--28 is the range over three seeds.}
    \label{tab:sources}
    \begin{tabularx}{\linewidth}{X c}
    \toprule
    \textbf{Source} & \textbf{Cases solved} \\
    \midrule
    Random search, 2 attempts & 26--28 \\
    Random search, 32 attempts & 54 \\
    Contact-GraspNet, top-40 grasps & 56 \\
    Human in VR (robust) & 58 \\
    \bottomrule
    \end{tabularx}
    \vspace{-2ex}
\end{table}

\subsection{Exploratory: Editing the Teacher's Weights}\label{sec:wsp}
Because Contact-GraspNet supplies the warm-start labels, a better teacher could improve every later stage. We tested whether the editing tools of IWIBench~\cite{iwibench} can improve it directly, without gradients or training. Following weight-space programming~\cite{wsp_blog}, a language model (Claude Opus 5.5) read and edited the network's 2.2M parameters through IWIBench's tensor tools and measured each candidate only on practice objects: one instance of each of the 57 training families, executing the top-ranked grasp at seeds separate from the test seeds (116 flat and 116 tilted episodes; two objects were executed three times). The test families and seeds were withheld from it. In IWIBench's terms the objective was stimulatory: to make the teacher rank grasps better than it had learned to.

The model found a structural property of the released PyTorch checkpoint. The confidence head ends in a ReLU followed by a linear layer whose 128 weights are all negative ($-0.46$ to $-0.04$) and whose bias is $-0.877$. Because the ReLU outputs are non-negative, which we confirmed on about 475k points from all practice objects (minimum exactly 0), no grasp can score above $\sigma(-0.877) = 0.294$, and a grasp reaches this value only when all 128 inputs are zero (1.3\% of points). These points produce ties. On 25 practice objects the top-ranked grasp is tied with 1 to 199 others, so the executed grasp is decided by the label generator's ordering rather than by the network. On 25 objects the scores are graded, with no tie at the top. On the remaining 7, no grasp clears Contact-GraspNet's absolute confidence threshold of 0.15, and the simulator falls back to a fixed default pinch. Flat practice episodes lifted 44 of 51 times on tied objects, 37 of 51 on graded ones, and 4 of 14 on unlabelled ones (85 of 116 overall; tilted 32/51, 20/51, and 3/14, 55 of 116). The groups contain different objects, so this comparison is not causal.

The model tried two edits, both modulatory in IWIBench's terms, since they unblock an existing capability rather than add one. Shifting a normalization bias inside the head broke almost all ties (one object remained tied) and changed the top-ranked grasp on 47 objects. Raising the output bias alone (to 1.623) preserves the ranking of every point, but because the threshold is absolute it lets about 2.6 times as many points pass (109k to 289k), gives grasps to all 7 unlabelled objects, and, through Contact-GraspNet's farthest-point sampling of candidates above the threshold, changes the top-ranked grasp on 20 objects. Both edits lowered practice success (flat 85 to 79 and 83; tilted 55 to 53 and 51), so the model reverted them and submitted the unedited weights. Under the bias edit, the flat loss came from objects with graded scores (four lifts lost, none gained), while the newly labelled objects were unchanged at 4 of 14. The result is negative: lifting the ceiling did not improve the teacher. It is consistent with the observation that weight edits can remove an obstacle that hides an ability but cannot add one~\cite{wsp_blog}; here the obstacle was not limiting performance.

The run also exposed variance in our Contact-GraspNet baseline. Regenerating its labels with the same weights and a different random camera viewpoint gave 64.0\% flat and 39.5\% tilted on 200 episodes each at the same evaluation seeds, against 55.1\% and 50.3\% in Table~\ref{tab:main}. The Contact-GraspNet rows should be read as one draw from a spread of roughly ten points.

\subsection{Limitations}\label{sec:limits}
\emph{Case selection.} Hard cases were chosen because the final RL-alone policies failed them, so RL alone is scored on cases selected against it, while the warm start and RL + demos are not. This regression effect favors both over RL alone. RL alone does score above the selection threshold, and the two conditions are level on hidden cases at iteration 25, which suggests that this does not explain the whole gap. A fresh RL-alone seed that played no part in selection measures it: it scores 21.2\% on demonstrated and 26.1\% on hidden cases. The demonstrated-case gain therefore does not depend on selection (45.2\% vs.\ 21.2\%), while part of the hidden-case gain may (33.6\% vs.\ 26.1\%, against 23.2\% for the selecting seeds).

\emph{Confounded components.} RL + demos differs from RL alone in targeted replay of the hard cases, contact supervision, and imitation at the same time, so our results do not isolate the human contribution. Matched RL-alone and RL + demos runs, together with an automated-supervision condition that replays the same cases with machine-found grasps under the same recipe and budget, would separate these effects.

\emph{Generalization.} The split is by case, stratified within family. Of the 71 hidden cases, 44 use an object instance that was also demonstrated, 25 use another instance of a demonstrated family, and two use a family with no demonstrations. The hidden-case gain therefore shows transfer to new situations on similar objects, not to new object categories. No hard case comes from a test family, and test objects show no significant gain.

\emph{Statistical power.} Case-level tests treat cases as the unit and do not capture variation between training runs. Hidden-case scores vary from 26.1\% to 40.8\% across RL + demos seeds, against 22.5\% to 24.6\% for RL alone. Every RL + demos seed exceeds every RL-alone seed at iteration 150, but with three seeds per condition an exact seed-level test cannot go below p = 0.1. The p-values are also not corrected for the multiple comparisons reported.

\emph{Object size and observation coverage.} The policy targets objects that fit the parallel-jaw gripper: grasp contacts lie within 12\,cm of the observed centroid and the jaw opens 8\,cm, so it needs a part of the object narrower than the opening. It was also trained on point clouds that cover most of the object surface. A single depth view of one side shifts the observed centroid and hides the far side, and we expect lower success from such partial views; a robot can meet this requirement by fusing several views, for example from a wrist camera. Training on rendered partial views, which our environment supports but our runs did not use, and grasping objects larger than the gripper's range are left to future work.

\emph{Baseline variance.} The Contact-GraspNet rows of Table~\ref{tab:main} come from one labelling run; a second camera viewpoint moved them by about ten points (Sec.~\ref{sec:wsp}). The warm start imitates these labels, so its scores may vary similarly. 

\emph{Contact fidelity.} Every object uses the same mass (0.2\,kg) and friction, objects are held still until the fingers close, and the contact model was not calibrated against a real gripper. A policy can exploit such a model with grasps that succeed in simulation but do not transfer~\cite{tutor2026ti01}.

\emph{Scope.} All results come from simulation, a floating hand, one demonstrator who is an author, and one demonstration round. The third RL + demos seed was interrupted by a machine suspend and restarted from scratch with the same recipe. The demonstrated-case score declines slightly as training continues, from 52.5\% at iteration 25 to 45.2\% at iteration 150, so longer training may erode part of the benefit.

\section{Conclusion}\label{sec:con}

We presented a failure loop in which persistent RL grasp failures are sent to VR, a person shows two grasp points, and verified demonstrations are used during a retraining run from the same warm start. RL improved on its Contact-GraspNet starting point for unseen objects (83.1\% vs. 55.1\% flat), but lost specific hard cases. With 58 demonstrations from a 32-minute session, RL + demos nearly doubled success on the demonstrated hard cases compared to RL alone, raised success on undemonstrated poses of the demonstrated objects by about ten points, back to the warm start's level, and showed no measurable loss on held-out test objects. We found no evidence of faster learning, and automated grasp search also solves many of the same cases. Next steps are matched RL-alone and RL + demos runs with an automated-supervision condition under the same recipe and budget, to isolate the human contribution, more demonstrators and seeds, training on partial camera views so that a single viewpoint suffices, and execution on a real robot.

\section*{Acknowledgment}
AI tools (Claude Opus 5.5, Google Photos) were used to edit the text of this manuscript for correctness and to edit images. The authors reviewed all AI-assisted edits and take full responsibility for the content of this paper.

\bibliographystyle{IEEEtran}
\bibliography{references}

\end{document}